\RequirePackage{fix-cm}
\documentclass[twocolumn]{svjour3}          % twocolumn
\smartqed  % flush right qed marks, e.g. at end of proof
\usepackage{graphicx}

\usepackage{url}
\usepackage{algorithmic}
\usepackage{amsmath}
\usepackage{amssymb}
\usepackage{multirow}
\usepackage{booktabs}
\usepackage{color}
\usepackage{setspace}
\usepackage{tabularx}
\usepackage{textcomp}
\usepackage{caption}
\usepackage{subcaption}
\usepackage[font=large]{subcaption}
\usepackage{emptypage}
\usepackage{float}
\usepackage[hidelinks,breaklinks]{hyperref}
\usepackage{multirow}
\usepackage{amsmath}
\usepackage{comment}

\def\makeheadbox{\relax}
\journalname{Journal of Real-Time Image Processing}
\begin{document}

\title{Efficient Vision-based Vehicle Speed Estimation%\thanks{Grants or other notes
}


\author{Andrej Macko        \and
        Lukáš Gajdošech     \and
        Viktor Kocur%etc.
}


\institute{A. Macko \at
              Photoneo, s.r.o., Plyn\'arensk\'a 6, Bratislava \\
              % Tel.: +123-45-678910\\
              \email{macko@photoneo.com}           %  \\
           \and
           L. Gajdo\v{s}ech, V. Kocur \at
           Faculty of Mathematics, Physics and Informatics, Comenius University Bratislava \\
           \email{\{lukas.gajdosech,viktor.kocur\}@fmph.uniba.sk}
}

\date{~}

\maketitle

\begin{abstract}
This paper presents a computationally efficient method for vehicle speed estimation from traffic camera footage. Building upon previous work that utilizes 3D bounding boxes derived from 2D detections and vanishing point geometry, we introduce several improvements to enhance real-time performance. We evaluate our method in several variants on the BrnoCompSpeed dataset in terms of vehicle detection and speed estimation accuracy. Our extensive evaluation across various hardware platforms, including edge devices, demonstrates significant gains in frames per second (FPS) compared to the prior state-of-the-art, while maintaining comparable or improved speed estimation accuracy.  We analyze the trade-off between accuracy and computational cost, showing that smaller models utilizing post-training quantization offer the best balance for real-world deployment. Our best performing model beats previous state-of-the-art in terms of median vehicle speed estimation error (0.58 km/h vs. 0.60 km/h), detection precision (91.02\% vs 87.08\%) and recall (91.14\% vs. 83.32\%) while also being 5.5 times faster.
\end{abstract}

%\section{Introduction}

\section{Introduction}
In today's fast-paced, urbanized world, intelligent transportation systems have become increasingly important in managing the growing complexities of transportation networks. Accurately measuring vehicle speeds is crucial for effective traffic management, enforcing speed limits, and developing intelligent transportation systems. Consequently, there has been growing interest in applying computer vision and machine learning techniques to develop reliable and efficient traffic analysis systems~\cite{chen2022review}.

Processing the vast amount of real-time visual data cameras generate poses significant computational challenges since current vision-based traffic surveillance methods rely on deep learning. These computational demands can be addressed in two main ways. The computations can be performed in cloud, where server-grade hardware can be utilized for efficient computation. The disadvantage of this approach is the need for high bandwidth and complex network infrastructure. As an alternative, edge computing processes sensor data closer to where the data are generated, thereby balancing the computing load and saving network resources. At the same time, edge computing has the potential for improved privacy protection by not transmitting all the raw data to the cloud datacenters~\cite{zhou2021intelligent}. 

When considering vision-based vehicle speed estimation within the edge computing paradigm it is important to consider the capabilities of edge computing hardware in design of the methods. Therefore, in this paper we propose modifications to the state-of-the-art vehicle speed estimation method~\cite{kocur2020detection} with focus on increased computational efficiency. We train several vehicle detection models with varying sizes and evaluate them using six different HW systems with focus on edge computing devices in terms of vehicle speed measurement accuracy and computational costs on the BrnoCompSpeed dataset~\cite{sochor2018comprehensive}. We assess the impact of operational precision, model size and image input size on both accuracy and computational efficiency of the networks. We show that the improved accuracy in terms of 2D bounding box localization for larger models does not necessarily translate to improvements in terms of vehicle speed measurement thus making smaller models a clear choice for vision-based vehicle speed estimation on edge.

The evaluation shows that our best model performs better than the previous state-of-the-art method~\cite{kocur2020detection} in terms of median vehicle speed estimation error (0.58 km/h vs. 0.60 km/h), detection precision (91.02\% vs 87.08\%) and recall (91.14\% vs. 83.32\%) while also being 5.5 times faster. We make our code and the trained models publicly available.\footnote{\url{https://github.com/gajdosech2/Vehicle-Speed-Estimation-YOLOv6-3D}}
\section{Related Works}
%In the Related Works chapter, we will delve into the dataset utilized for our study, the foundational method for speed measurement that we will be expanding upon, and an overview of the current leading object detector in the field. By examining these components, we will establish a comprehensive understanding of the existing landscape and identify areas where our research can contribute to advancements in object detection and speed measurement techniques.

Vision-based vehicle speed estimation is broadly composed of several steps: traffic camera calibration, vehicle detection and tracking. In this section we present an overview of previous literature for the individual tasks and overall vehicle speed estimation systems.

\subsection{Traffic Camera Calibration}

Traffic camera calibration enables accurate measurements of distances in the road plane by estimating camera intrinsics and extrinsics. Traffic cameras can be calibrated manually using a calibration pattern~\cite{hepattern} or distances measured in the road plane~\cite{luvizon2014}. Several semi-automatic methods combine automated approaches with at least one known metric distance in the scene based on vanishing point detection~\cite{maduro,you,zhang2013,kocur2021traffic,li2023automatic} or parallel curves~\cite{parallelcurves}.

Fully-automatic methods do not require additional metric information as they recover the scale automatically. Some methods are based on detection of vanishing points and estimation of the scale based on vehicle dimensions~\cite{dubska2014,sochor2017traffic}. Other methods rely on detecting landmarks of vehicles~\cite{filipiak,autocalib,landmarkcalib} or their bounding boxes and assuming their known shape~\cite{revaud2021robust}. \cite{vuong2024toward} relies on Google Street View to produce a metric 3D model of the traffic scene which can be used to calibrate the camera.

\subsection{Object Detection}

Object detection is a key component of vison-based vehicle speed estimation. In recent years this task was dominated by deep learning approaches. Various detection frameworks have been proposed based on bounding box anchors in single~\cite{lin2017focal} or two stage processing~\cite{ren2015faster}. More recently anchor-free approaches~\cite{Zhou2019,tian2019fcos} have emerged including approaches based on transformers~\cite{carion2020}.

More recently, improvements in object detectors were focused on improved computational efficiency. The YOLO series~\cite{redmon2016you} represents a staple among efficient object detectors. It has been refined continuously over the past few years~\cite{terven2023}, with YOLOv4~\cite{bochkovskiy2020yolov4} introducing CSP-based backbones and PANet for improved feature aggregation and faster inference. YOLOv5~\cite{yolov5} further streamlined the pipeline in a modular framework with multiple scale variants that cater to various resource constraints. YOLOv6~\cite{li2022yolov6,li2023yolov6} then built on these improvements by incorporating reparameterized backbones (e.g., EfficientRep) and decoupled detection heads specifically optimized for low-power, edge deployments. In parallel, other YOLO variants have been developed by utilizing various techniques~\cite{sapkota2025,Xu2022PP} and neural architecture search~\cite{Xu2022DAMO} to push the speed-accuracy tradeoff even further. Alternatives to YOLO~\cite{lyu2022RTMDet,Tan2019} also provide object detectors focused on efficiency in low-compute settings.

The performance of object detectors can be further increased with techniques of quantization and knowledge distillation. Quantization-aware training (QAT) and post-training quantization (PTQ) are the two possible approaches to quantization. QAT integrates quantization during training, allowing the model to adjust and mitigate accuracy loss, while PTQ applies quantization after training using calibration data to determine scaling factors and clipping ranges \cite{gholami2021survey}. The second technique, knowledge distillation, transfers knowledge from a large model to a smaller model by minimizing the divergence between their outputs \cite{basterrech2022tracking}, ensuring efficiency while maintaining performance \cite{li2023kd,li2022yolov6}. Post‑training quantization strategies have been successfully applied in other real‑time detection studies on embedded platforms \cite{lazarevich2023}.

\subsection{Tracking}

Tracking is necessary to associate individual vehicle detections across multiple frames. Common tracking methods include the Kalman filter~\cite{kalman}. SORT~\cite{sort} and its many variants~\cite{deepsort,Cao23,du2023strongsort} use Kalman filter for tracking object bounding boxes. Since modern object detectors provide good accuracy it is also possible to rely on a simple IOU based tracker~\cite{bochinski2017high}. An alternative approach is to perform tracking and detection jointly in a single neural network~\cite{zhou2020tracking,tracktor}. 

\subsection{Vision-based Vehicle Speed Estimation}

Several pipelines for vehicle speed estimation have been proposed. Most pipelines are composed of camera calibration, vehicle detection and tracking with the final speed obtained from the vehicle positions within the track~\cite{zhang2022monocular}. Vehicle speed estimation was the focus of one of the tracks of AI City Challenge 2018~\cite{aicc2018}. The challenge participants used deep learning-based object detectors in combination with various semi-automatic methods for camera calibration. To obtain the final speeds the participants used medians, means, percentiles of inter-frame distances or their combinations.

In addition to the standard 2D bounding boxes, \cite{dubska2014,sochor2017traffic} detect masks of vehicles to construct 3D bounding boxes based on known vanishing points. In \cite{kocur2020detection,kocur2019perspective} 3D bounding boxes are directly estimated by first rectifying the scene based on the known vanishing points and then using a modified 2D object detector which outputs one additional parameter to provide a 3D bounding box. Direct regression of the 3D bounding box via its centroid and vertices is also possible using a specialized neural network~\cite{tang2023centerloc3d}. Some works~\cite{filipiak,luvizon2017} detect license plates instead of the full vehicles to estimate vehicle speeds.

In a completely orthogonal approach, authors of \cite{barros2021deep} propose a deep neural network which estimates speeds of vehicles directly without previous camera calibration.

\subsection{Evaluation Datasets}

Evaluating vision-based speed estimation methods requires datasets with accurate vehicle ground truth speeds. 
% Many methods are evaluated on datasets with inaccurate ground truth \cite{pattern} or small number of ground truth annotations \cite{dubska2014, maduro, speed1, luvizon2014}.
To provide such data the authors of BrnoCompSpeed dataset \cite{sochor2018comprehensive} captured 21 hour-long videos taken from a surveillance viewpoint above various roads in the city of Brno. The videos were recorded in 7 different locations. Every road section was recorded using three cameras from different viewing angles. To record the speed of vehicles passing through the surveillance viewpoint, the authors used GPS-synchronized LIDAR thus providing accurate speed estimates.

A smaller dataset was published by Luviz\'{o}n et al. \cite{luvizon2017} comprising of 5 hours of video of a small section of a three lane road leading up to an intersection. Ground truth speeds of vehicles are obtained via induction loops installed in the road. This dataset also contains ground truth annotations of vehicle license plates.

The 2018 NVIDIA AI City Challenge \cite{aicc2018} included a track for speed estimation from video footage. To evaluate the challenge the organizers collected a dataset of 27 HD videos, each one minute long. Unfortunately, the vehicle speed estimation annotations were not made available publicly, instead requiring challenge participants to use an evaluation server.

\section{Computationally Efficient Vehicle Speed Estimation}
\begin{figure}
    \centering
    \includegraphics[width=\linewidth]{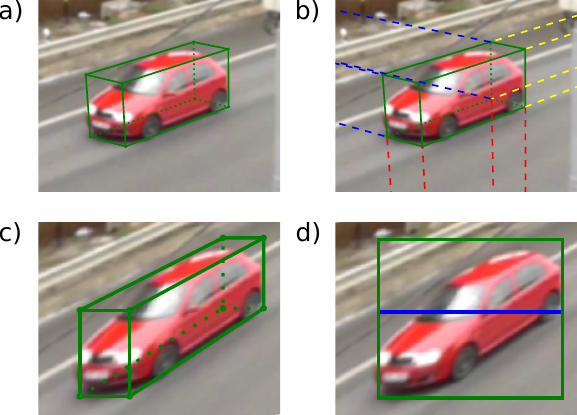}
    \caption{When vehicles travel on a straight road their 3D bounding box (a) is aligned with three relevant vanishing points (b). Knowledge of the vanishing points positions, which can be obtained by automatically calibrating the camera (e.g. using~\cite{sochor2017traffic}), can be used to rectify the image (c). In the rectified image the task of estimating the 3D bounding box is reduced to finding a 2D bounding box with one additional parameter $c_c$ which determines the position of the top frontal edge (blue) of the 3D bounding box in the 2D bounding box (green). Image adopted from~\cite{kocur2020detection}.}
    \label{fig:mvaa}
\end{figure}

In this section we provide details about the proposed efficient speed estimation method. The method is based on~\cite{kocur2020detection} with several modifications that result in significantly better computational efficiency compared to the baseline. To make the paper self-contained we will first provide a brief overview of the method presented in~\cite{kocur2020detection} and later provide information on our proposed modifications which result in significantly better computational efficiency.

\subsection{Baseline Method}

The baseline method~\cite{kocur2020detection} presents a pipeline for vehicle speed estimation by detecting 3D bounding boxes of vehicles. Unlike traditional 2D bounding boxes, the 3D bounding boxes provide a consistent tracking point at the center of the bottom frontal edge, ensuring reliable speed measurement regardless of the camera angle. This leads to significant improvements in terms of vehicle speed measurement accuracy over a naive approach that relies on 2D bounding boxes.

% TODO figure 
The pipeline consists of several key steps: camera calibration, image transformation, vehicle detection, 3D bounding box reconstruction, vehicle tracking, and speed estimation. In the first step the traffic camera is calibrated using~\cite{sochor2017traffic}. This calibration method detects vanishing points relevant to the scene and also estimates the scale enabling for accurate metric measurements in the road plane. Based on the obtained vanishing points a perspective image transformation is constructed such that the directions corresponding with 2 of the 3 relevant vanishing points are aligned with image axes. This rectifies the image enabling for easier detection and at the same times makes it straightforward to parametrize a 3D bounding box of a vehicle as a 2D bounding box with one additional parameter. For a visual representation of this process see Figure~\ref{fig:mvaa}. The 2D bounding boxes with the additional parameter are detected using a modified RetinaNet~\cite{lin2017focal}. After detection the 3D bounding boxes are constructed in the original frame based on the known positions of the vanishing point. Finally, the vehicles are tracked using a simple IOU tracker~\cite{bochinski2017high} and their speed is estimated by calculating the median distance traveled between individual frames.

\subsection{Improved Base Detector}

To improve the computational efficiency of the baseline method~\cite{kocur2020detection} we propose to use YOLOv6 v3.0~\cite{li2023yolov6} instead of RetinaNet~\cite{lin2017focal}. YOLOv6 is an anchor-free object detector based on a point-based paradigm \cite{tian2019fcos}. The network predicts a 4D output containing the distributions of distances of the object bounding box edges from its center point, classification and objectness output. This output is then converted into final bounding boxes using non-maximum suppression.

The key component of YOLOv6 for fast inference is the RepVGG architecture \cite{ding2021repvgg}, which leverages structural re-parameterization to optimize performance. During training, RepVGG incorporates multi-branch structures inspired by ResNet \cite{he2016deep}, including identity and $1 \times 1$
branches. After training, these branches are transformed into a single $3\times 3$ convolutional layer using algebraic operations, combining the parameters of the original branches and batch normalization \cite{ioffe2015batch}. This results in an inference-time model composed solely of $3 \times 3$ convolutions with ReLU, making RepVGG highly efficient on GPUs. YOLOv6 is available in four model sizes: Nano, Small, Medium and Large.

While newer variants such as YOLOv7, YOLOv8, and YOLO‑NAS achieve incremental accuracy gains through additional complexities in network design and training strategies \cite{terven2023,sapkota2025}, their increased computational overhead and implementation intricacies make them less suitable for our focus on deployment rather than absolute performance. Several studies have demonstrated that, for applications where real‑time performance and ease of integration are prioritized over peak accuracy, models like YOLOv6 deliver competitive detection capabilities with a favorable accuracy–latency trade‑off on embedded platforms \cite{lazarevich2023,ling2024}. Our goal is to systematically examine the effects of different operational configurations, their influence is expected to be largely independent of the specific detector architecture. Additionally, our chosen revision 3.0 of the YOLOv6 includes improvements that put it ahead of newer versions even in raw accuracy. 

% \begin{figure}
%     \centering
%     \includegraphics[width=1\columnwidth]{images/efficienthead_modified_v2.png}
%     \caption{The architecture of modified YOLOv6 efficient head \cite{li2022yolov6}. Added blocks and layers are highlighted with color. The convolution block is denoted with a blue rectangle, and the convolution layer is denoted with an orange rectangle.}
%     \label{img:modified_eff_head}
% \end{figure}

The standard YOLOv6 architecture produces output classes and 2D bounding boxes of objects. To include the additional parameter $c_c$ introduced in~\cite{kocur2020detection} (see Figure~\ref{fig:mvaa}), we added one convolutional block and one convolution layer for the final prediction to the efficient head. %This modification can be seen in Figure \ref{img:modified_eff_head}.
First, the convolutional block comprises a convolution layer with kernel size $3 \times 3$, batch normalization \cite{ioffe2015batch}, and SiLU \cite{elfwing2018sigmoid} activation function. Output of this convolution block is passed to the convolution layer with kernel size $1 \times 1$, and with output size one as we predict the $c_c$ parameter.

% YOLOv6 uses two types of loss functions. The first one is VariFocal Loss (VFL) \cite{zhang2021varifocalnet} for classification loss, and the second one is SIoU \cite{gevorgyan2022siou} (for nano and small model) or GIoU \cite{rezatofighi2019generalized} (medium and large).

\subsection{Training Data}\label{sec:data_ann}

To train our model, we used two datasets BrnoCompSpeed \cite{sochor2018comprehensive} and BoxCar116 \cite{Sochor2018}. The BrnoCompSpeed dataset \cite{sochor2018comprehensive} contains has 21 videos that were recorded in 7 different sessions. We use the first four sessions from the BrnoCompSpeed dataset to training and validation. To obtain 3D bounding box annotations we follow the procedure from~\cite{kocur2020detection} which combines vehicle masks obtained using~\cite{he2017mask} and camera calibration data provided with the dataset. This procedure directly provides transformed images and annotations in the form of a 2D bounding box with the additional parameter $c_c$ (see Figure~\ref{fig:mvaa}). We also use BoxCars116k~\cite{sochor2018comprehensive} which contains images of individual vehicles along with camera calibration information and 3D bounding boxes. We use this information to obtain annotations and transformed images.

\subsection{Training}

\begin{figure}
    \includegraphics[width=0.32\columnwidth]{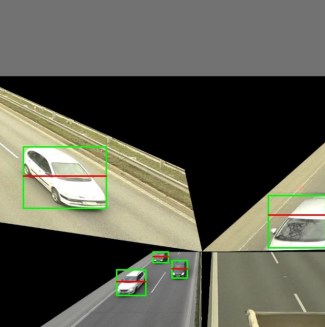} \hfill    
    \includegraphics[width=0.32\columnwidth]{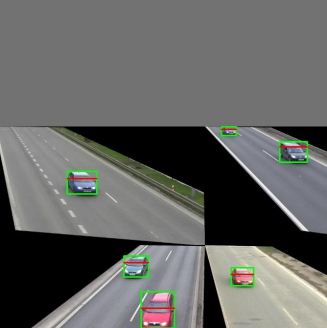} \hfill    
    \includegraphics[width=0.32\columnwidth]{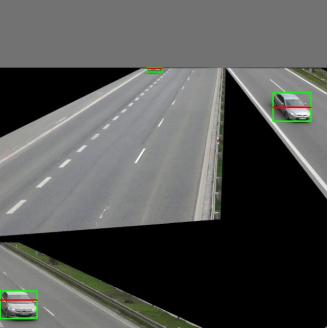}
    \caption{Example of mosaic data augmentation with their respective annotations adjusted to match the new position.}
    \label{img:mosaic}
\end{figure}

During the training process instead of preprocessing proposed in~\cite{kocur2020detection}, we applied mosaic data augmentation. In mosaic augmentation, four images are randomly chosen from the training dataset and combined into a single image by placing them in a 2x2 grid. The new image contains objects from all four original images. Images are additionally flipped and scaled randomly. Their respective annotations (bounding boxes and parameter $c_c$) are adjusted to match the new positions. Examples of augmented data are shown in Figure \ref{img:mosaic}.

%One problem with this augment method is that, in some cases, when images are rotated, scaled, and cropped randomly, they will not contain any vehicle. In these cases, we undo changes and do it again.

%We trained four models of different sizes: nano, small, medium, and large. Nano and small have the same architecture (EfficientRep, RepBiFPANNeck, Efficient decoupled head), but they have different numbers of parameters. Similarly, the medium and the large architectures have the same architectures design except for a different number of parameters due to channel. \TODO{Vysvetlit}

We trained all four available model sizes of the YOLOv6 architecture. During training ground truth 2D bounding boxes are assigned to network outputs using~\cite{feng2021tood}. We incorporate regression of the parameter $c_c$ into the overall loss function using the mean squared error:
\begin{equation}\label{eq:mse_cc}
L_{c} = \frac{1}{N}\sum_{i=1}^N(c_{c,i}^p - c_{c,i}^g)^2.
\end{equation}
where $c_{c, i}^p$ is predicted $c_c$ parameter, and $c_{c, i}^g$ is the ground truth value of $i$-th of the $N$ assigned bounding boxes.

The overall loss is calculated as
\begin{equation}
L_{total} = L_{cls} + L_{iou} + L_{c},
\end{equation}
where $L_{cls}$ is the VariFocal classification loss \cite{zhang2021varifocalnet}, $L_{iou}$ is SIoU \cite{gevorgyan2022siou} (for nano and small models) or GIoU \cite{rezatofighi2019generalized} (medium and large) for 2D bounding box regression and $L_{c}$ is the mean-squared error (\ref{eq:mse_cc}) for the $c_c$ parameter.

\begin{table}[h]
    \centering
    \resizebox{\linewidth}{!}{
    \begin{tabular}{@{}cccccc@{}}
    \toprule
    Model  & mAP 0.5:0.95 & mAR 0.5:0.95 & $c_c$ error & Params  & FLOPs    \\ \midrule
    Nano   & 45.0 \%      & 67.4 \%      & 0.0289      & 4.85 M  & 11.83 G  \\
    Small  & 51.0 \%      & 72.6 \%      & 0.0267      & 19.33 M & 46.86 G  \\
    Medium & 51.0 \%      & 71.4 \%      & 0.0219      & 36.62 M & 89.24 G  \\
    Large  & 52.0 \%      & 72.0 \%      & 0.0251      & 62.71 M & 156.36 G \\ \midrule
    Nano distill  & 46.0\%       & 67.0\%       & 0.0224      & 4.85 M  & 11.83 G \\
    Small distill & 50.5\%       & 68.9\%       & 0.0214      & 19.33 M & 46.86 G \\ \midrule
    Nano quantized  & 38.9\%       & 59.2\%       & 0.0297      & 4.85 M & - \\
    Small quantized & 44.3\%       & 65.1\%       & 0.0245      & 19.33 M & -\\ \bottomrule
    \end{tabular}
    }
    \caption{Evaluation of our modified YOLOv6 models the validation set (see Sec.~\ref{sec:data_ann}). mAP and mAR denote mean average precision and recall of the 2D bounding boxes respectively. $c_c$ denotes the mean of error~\eqref{eq:mse_cc} on the validation set.}
    \label{tb:vanilla_eval}
\end{table}

We trained all of our models for 30 epochs, after which there were no further improvements in terms of validation loss. The first three epochs were warm-up epochs. We chose Stochastic Gradient Descent (SGD) with a momentum = 0.937 as the optimizing algorithm, a learning rate of 0.01, and a cosine learning rate scheduler. After the full three epochs we selected a snapshot of the model which performed best on the validation set. Based on the superior validation results of the Large model we also utilized knowledge distillation~\cite{li2023kd} to distill it to the Nano and Small models~\cite{li2022yolov6}. We trained them for 30 epochs. The first three epochs were warm-up epochs. We chose Stochastic Gradient Descent (SGD) with a momentum = 0.937 as the optimizing algorithm, a learning rate of 0.01, and a cosine learning rate scheduler. The results for the trained models on the validation set are provided in Table~\ref{tb:vanilla_eval}. We note that distillation brought some improvement for the Nano model in terms of mAP, but the distilled counterpart of the Small model performs worse. 

\subsection{Post-Training Quantization}

For our chosen quantization approach, we opted for PTQ over QAT due to its lower computational cost and simpler deployment process, making it more practical for real-time traffic speed estimation while maintaining accuracy.
%Our post‑training quantization process follows the pipeline described in the official YOLOv6 repository by Meituan\footnote{https://github.com/meituan/YOLOv6}
It follows the pipeline suggested by YOLOv6 authors, where the floating‑point ONNX model is converted into a TensorRT v8 engine using its default calibration process \cite{li2022yolov6}.
%In this pipeline, TensorRT’s \texttt{IInt8EntropyCalibrator2} is employed to perform image calibration. 
The calibrator runs a representative set of images through the network to collect activation statistics and compute optimal scaling factors that map the FP32 dynamic range to the reduced INT8 space. This mechanism plays a crucial role in maintaining detection accuracy while significantly reducing inference latency and memory footprint.  

For the calibration process, we used training data described in Subsection~\ref{sec:data_ann} with a batch size of 32 across 32 calibration batches. This resulted in processing 1024 training images, allowing us to capture comprehensive statistical properties of activations across all network layers.

% We explained the quantization technique and its benefic in section \ref{sec:ptq}. Authors of YOLOv6 suggested switching RepVGG architecture \cite{ding2021repvgg} to QARepVGG (Quantization-aware RepVGG) \cite{chu2022make}. However, because of re-parametrization, doing PTQ lowers accuracy by 20 $\%$. So, for quantization models, RepVGG was a switch with QARepVGG.

\begin{table*}[h]
\centering
\resizebox{\linewidth}{!}{
\begin{tabular}{cccccccc}
~ & Model &
  \begin{tabular}[c]{@{}c@{}}Input size\\ (px)\end{tabular} &
  \begin{tabular}[c]{@{}c@{}}Mean error \\ (km/h)\end{tabular} &
  \begin{tabular}[c]{@{}c@{}}Median error \\ (km/h)\end{tabular} &
  \begin{tabular}[c]{@{}c@{}}95-th percentile \\ (km/h)\end{tabular} &
  \begin{tabular}[c]{@{}c@{}}Mean precision \\ (\%)\end{tabular} &
  \begin{tabular}[c]{@{}c@{}}Mean recall \\ (\%)\end{tabular} \\ \cline{2-8}

\multirow{24}{*}{\rotatebox{90}{F32}} & Dubská et al.~\cite{dubska2014}                & -                           & 8.22  & 7.87  & 10.43 & 73.48 & 90.08 \\ \cline{2-8}
& SochorAuto~\cite{sochor2017traffic}     & -                           & 1.10  & 0.97  & 2.22  & 90.72 & 83.34           \\ \cline{2-8}
& SochorManual~\cite{sochor2017traffic}         & -                           & 1.04  & 0.83  & 2.35  & 90.72 & 83.34   \\ \cline{2-8}
& Learned+RANSAC~\cite{revaud2021robust} & - & 2.15 & 1.60 & - & - & - \\ \cline{2-8}

 & \multirow{3}{*}{Transform3D~\cite{kocur2020detection}} & 480 x 270                   & 0.92  & 0.72  & 2.35  & 89.26 & 79.99  \\ 
&                                       & 640 x 360                   & \textbf{0.79}  & 0.60  & \textbf{1.96}  & 87.08 & 83.32  \\ 
&                                       & 540 x 960                   & 1.09  & 0.84  & 2.65  & 88.06 & 85.30  \\ \cmidrule{2-8}

& \multirow{3}{*}{Nano}          & 480 x 270 & 0.87          & 0.69          & 2.09          & 88.08          & 91.23           \\
&                               & 640 x 360 & 0.80 & 0.64          & 2.04          & 89.69          & 92.25           \\
&                               & 960 x 540 & 0.82          & 0.66          & 2.17          & 91.15          & 90.10           \\ \cline{2-8}
& \multirow{3}{*}{Nano distill}  & 480 x 270 & 0.98          & 0.74          & 2.47          & 81.71          & 87.39           \\
&                               & 640 x 360 & 0.90          & 0.81          & 2.24          & 86.00          & 89.15           \\
&                               & 960 x 540 & 0.87          & 0.70          & 2.19          & 84.72          & 86.53           \\ \cline{2-8}
& \multirow{3}{*}{Small}         & 480 x 270 & 0.87          & 0.68          & 2.16          & 91.08          & \textbf{92.85}  \\
&                               & 640 x 360 & 0.81          & \textbf{0.58} & 2.01 & 91.02          & 92.16           \\
&                               & 960 x 540 & 0.81          & 0.63          & 2.08          & \textbf{92.11} & 91.14           \\ \cline{2-8}
& \multirow{3}{*}{Small distill} & 480 x 270 & 0.91          & 0.73          & 2.24          & 85.45          & 91.00           \\
&                               & 640 x 360 & 0.84          & 0.65          & 2.16          & 85.05          & 89.58           \\
&                               & 960 x 540 & 0.86          & 0.67          & 2.16          & 84.73          & 90.07           \\ \cline{2-8}
& \multirow{3}{*}{Medium}        & 480 x 270 & 0.84          & 0.67          & 2.15          & 91.55          & 92.44           \\
&                               & 640 x 360 & 0.86          & 0.69          & 2.21          & 91.32          & 90.68           \\
&                               & 960 x 540 & 0.83          & 0.66          & 2.14          & 91.10          & 91.00           \\ \cline{2-8}
& \multirow{3}{*}{Large}         & 480 x 270 & 0.87          & 0.70          & 2.25          & 90.85          & 91.12           \\
&                               & 640 x 360 & 0.84          & 0.66          & 2.16          & 90.34          & 90.67           \\
&                               & 960 x 540 & 0.82          & 0.64          & 2.13          & 91.21          & 90.85           \\ \hline
\multirow{6}{*}{\rotatebox{90}{INT8}} & \multirow{3}{*}{Nano}  & 480 x 270   & 0.89 & 0.69 & 2.15 & 88.05 & 91.51 \\
&                       & 640 x 360   & 0.82 & 0.65 & 2.19 & 89.62 & 91.27 \\
&                       & 960 x 540   & 0.87 & 0.69 & 2.14 & 90.92 & 90.13 \\ \cline{2-8}
& \multirow{3}{*}{Small} & 480 x 270   & 0.88 & 0.70 & 2.09 & 88.08 & 91.23 \\
&                       & 640 x 360   & 0.78 & 0.60 & 2.04 & 91.02 & 91.27 \\
&                       & 960 x 540   & 0.83 & 0.62 & 2.09 & 90.60 & 90.07 \\ \cline{2-8}
                               
\end{tabular}
}
\caption{Speed measurement evaluation on BrnoCompSpeed test split C~\cite{sochor2018comprehensive} with the exception of Learned+RANSAC~\cite{revaud2021robust} which was evaluated on the full dataset. The first column indicates operational precision for full precision floats (F32) and quantized models (INT8). Results for 16-bit floats are provided in the supplementary information.}
\label{tb:speed_acc}
\end{table*}

\section{Results}

\begin{table*}[h]
\centering
\small
% \resizebox{0.8\linewidth}{!}{
\begin{tabular}{@{}ccccccccc@{}}
\toprule
GPU &
  \begin{tabular}[c]{@{}c@{}}Memory\\ (GB)\end{tabular} &
  \begin{tabular}[c]{@{}c@{}}CUDA \\ cores\end{tabular} &
  \begin{tabular}[c]{@{}c@{}}Tensor \\ cores\end{tabular} &
  \begin{tabular}[c]{@{}c@{}}FP32\\ (TFLOPS)\end{tabular} &
  \begin{tabular}[c]{@{}c@{}}FP16\\ (TFLOPS)\end{tabular} &
  \begin{tabular}[c]{@{}c@{}}INT8\\ (TOPS)\end{tabular} &
  \begin{tabular}[c]{@{}c@{}}Release \\ year\end{tabular} & \begin{tabular}[c]{@{}c@{}}System \\ CPU\end{tabular} \\ \midrule
Titan V  & 12 & 5120 & 640 & 13.80 & 27.60 & 55.20 & 2017 & AMD Ryzen 7 2700 \\
RTX 2080 & 8  & 2944 & 368 & 10.07 & 20.14 & 41.20 & 2018 & AMD Ryzen 7 3700X \\
940M     & 2  & 384  & -   & 0.93  & 1.83  & -     & 2015 & Intel i5 6200U \\ 
Jetson TX2    & 4 & 256 & - & 0.66 & 1.33 & - & 2016 & - \\
Jetson Xavier & 8  & 384 & - & 1.41 & 2.82 & 21.00 & 2018 & -  \\
Jetson Orin   & 8  & 1024  & 32   & 1.56  & 3.13  & 70.00     & 2023 & - \\ \bottomrule
\end{tabular}
% }
\caption{Technical parameters of HW systems used in our computational efficiency evaluation.}
\label{tb:gpu_comperisons}
\end{table*}

In this section we present an evaluation of the trained models with respect to vehicle speed measurement accuracy. We also perform extensive evaluation of the various trained models in terms of computational efficiency.

\subsection{Speed Measurement Evaluation}
\label{subsec:speed_eval}

For vehicle speed measurement evaluation, we use the official evaluation tool\footnote{https://github.com/JakubSochor/BrnoCompSpeed} from the authors of \cite{sochor2018comprehensive}. The speed accuracy results for split C of the dataset are shown in Table~\ref{tb:speed_acc}. We provide results for the proposed methods in half precision (F16) in the supplementary information. Among the variants of our proposed method the Small models with 640x360 px input resulution performs the best in both full precision and quantized version. It performs on par with the previous state-of-the-art method~\cite{kocur2020detection} in terms of mean speed measurement error while achieving better median speed estimation error, and both detection recall and accuracy. In the next subsection we will show that in addition to achieving state-of-the-art vehicle speed estimation accuracy our method is also superior in terms of computational efficiency.

We also note an interesting observation that the larger models do not perform better than their smaller counterparts despite their better results in terms of mAP and mAR on the validation set (see Table~\ref{tb:vanilla_eval}). This may occur due to several reasons such as smaller models generalizing better. Another possibility is that the accuracy of the predicted bounding boxes in individual frames is not as important for the downstream task of vehicle speed estimation since the speed measurement is aggregated across multiple frames. We also note that increase in the size of the input image improves the speed measurement accuracy only up to the resolution of 640 x 360 which may be due to similar reasons.

\subsection{Computational Efficiency}

% \begin{table*}[h]
% \centering
% \begin{tabular}{@{}ccccc@{}}
% \toprule
%  & GPU & CPU & \begin{tabular}[c]{@{}c@{}}CPU \\ cores\end{tabular} & \begin{tabular}[c]{@{}c@{}}RAM\\ (GB)\end{tabular} \\ \midrule
% Machine 1 & Nvidia TITAN V  & \begin{tabular}[c]{@{}c@{}}AMD Ryzen 7 \\ 2700\end{tabular} & 8 & 48 \\
% Machine 2 & Nvidia RTX 2080 & \begin{tabular}[c]{@{}c@{}}AMD Ryzen 7\\ 3700X\end{tabular} & 8 & 32 \\
% Machine 3 & Nvidia 940M     & \begin{tabular}[c]{@{}c@{}}Intel I5 \\ 6200U\end{tabular}   & 2 & 8  \\ \bottomrule
% \end{tabular}
% \caption{Technical parameters of machines on which FPS was calculated.}
% \label{tb:machines}
% \end{table*}

For practical uses of vision-based vehicle speed estimation it is important to also consider the associated computational costs. We therefore perform an evaluation using a variety of HW systems including edge devices, consumer grade PCs and computational servers. The GPUs used for evaluation are listed in Table~\ref{tb:gpu_comperisons}.

\begin{table*}[]
\centering
\small
% \resizebox{0.8\linewidth}{!}{
\begin{tabular}{ccccccccc}
  Model &
  Data Type &
  \begin{tabular}[c]{@{}c@{}}Input size \\ (px)\end{tabular} &
  \begin{tabular}[c]{@{}c@{}}940M \\ (FPS)\end{tabular} &
  \begin{tabular}[c]{@{}c@{}}RTX 2080 \\ (FPS)\end{tabular} &
  \begin{tabular}[c]{@{}c@{}}Titan V \\ (FPS)\end{tabular} &
  \begin{tabular}[c]{@{}c@{}}TX2 \\ (FPS)\end{tabular} &
  \begin{tabular}[c]{@{}c@{}}Xavier \\ (FPS)\end{tabular} & 
  \begin{tabular}[c]{@{}c@{}}Orin \\ (FPS)\end{tabular} 
  \\ \hline
\multirow{3}{*}{Transform3D~\cite{kocur2020detection}} & \multirow{3}{*}{F32} & 480 x 270 & - & - & 70 & - & - & - \\
                                                       &                      & 640 x 360 & - & - & 62 & - & - & - \\
                                                       &                      & 960 x 540 & - & - & 43 & - & - & - \\
\hline
           &                       & 480 x 270 & - & 554 & 464 & - & 52 & 95 \\
           &                       & 640 x 360 & - & 355 & 355 & - & 38 & 68 \\
           & \multirow{-3}{*}{INT8}& 960 x 540 & - & 256 & 246 & - & 21 & 57 \\ 
           &                       & 480 x 270 & 49 & 436 & 357 & 19 & 44 & 76 \\
           &                       & 640 x 360 & 25 & 311 & 267 & 17 & 35 & 54 \\
           & \multirow{-3}{*}{FP16}& 960 x 540 & 14 & 194 & 179 & 9 & 27 & 38 \\
            &                       & 480 x 270 & 38 & 295 & 295 & 30 & 38 & 65 \\
            &                       & 640 x 360 & 20 & 224 & 221 & 19 & 29 & 43 \\
 
\multirow{-9}{*}{Nano}         & \multirow{-3}{*}{FP32}& 960 x 540 & 10 & 148 & 132 & 12 & 19 & 31 \\ 
\hline
%\multirow{9}{*}{Nano distill}  & \multirow{3}{*}{INT8} & 480 x 270 & - & - & - & - & - & - \\
%                               &                       & 640 x 360 & - & - & - & - & - & - \\
%                               &                       & 960 x 540 & - & - & - & - & - & - \\
%                               & \multirow{3}{*}{F16}  & 480 x 270 & ? & 360 & 357 & 32 & 45 & 76 \\
%                               &                       & 640 x 360 & ? & 317 & 261 & 18 & 35 & 54 \\
%                               &                       & 960 x 540 & ? & 196 & 135 & ? & 31 & 39 \\
%                               & \multirow{3}{*}{F32}  & 480 x 270 & ? & 296 & 289 & 35 & 39 & 65 \\
%                               &                       & 640 x 360 & ? & 227 & 217 & 19 & 30 & 46 \\
%                               &                       & 960 x 540 & ? & 145 & 140 & ? & 19 & 30 \\ \hline 
\multirow{9}{*}{Small}         & \multirow{3}{*}{INT8} & 480 x 270 & - & 489 & 392 & - & 44 & 126 \\
                               &                       & 640 x 360 & - & 302 & 302 & - & 33 & 90 \\
                               &                       & 960 x 540 & - & 205 & 198 & - & 16 & 95 \\
                               & \multirow{3}{*}{FP16} & 480 x 270 & 23 & 448 & 354 & 15 & 44 & 58 \\
                               &                       & 640 x 360 & 14 & 243 & 205 & 8 & 36 & 39 \\
                               &                       & 960 x 540 & 6 & 145 & 139 & 5 & 18 & 24 \\
                               & \multirow{3}{*}{FP32} & 480 x 270 & 17 & 229 & 223 & 17 & 28 & 45 \\
                               &                       & 640 x 360 & 11 & 142 & 142 & 9 & 22 & 30 \\
                               &                       & 960 x 540 & 4 & 95 & 98 & 5 & 10 & 18 \\ \hline
%\multirow{9}{*}{Small distill} & \multirow{3}{*}{INT8} & 480 x 270 & - & - & - & - & - & - \\
%                               &                       & 640 x 360 & - & - & - & - & - & - \\
%                               &                       & 960 x 540 & - & - & - & - & - & - \\
%                               & \multirow{3}{*}{F16}  & 480 x 270 & ? & 223 & 242 & 15 & 45 & 58 \\
%                               &                       & 640 x 360 & ? & 244 & 204 & ? & 38 & 38 \\
%                               &                       & 960 x 540 & ? & 144 & 170 & ? & 18 & 24 \\
%                               & \multirow{3}{*}{F32}  & 480 x 270 & ? & 226 & 219 & 18 & 27 & 46 \\
%                               &                       & 640 x 360 & ? & 143 & 143 & ? & 20 & 30 \\
%                               &                       & 960 x 540 & ? & 93 & 102 & ? & 10 & 17 \\ \hline 
\multirow{6}{*}{Medium}        & \multirow{3}{*}{FP16} & 480 x 270 & 13 & 356 & 330 & 8 & 36 & 39 \\
                               &                       & 640 x 360 & 6 & 183 & 174 & 5 & 21 & 24 \\
                               &                       & 960 x 540 & 2 & 110 & 108 & 2 & 12 & 14 \\
                               & \multirow{3}{*}{FP32} & 480 x 270 & 10 & 187 & 162 & 9 & 21 & 31 \\
                               &                       & 640 x 360 & 5 & 128 & 120 & 5 & 13 & 19 \\
                               &                       & 960 x 540 & 2 & 55 & 67 & 2 & 5 & 9 \\ \hline
\multirow{6}{*}{Large}         & \multirow{3}{*}{FP16} & 480 x 270 & - & 252 & 271 & - & 28 & 29 \\
                               &                       & 640 x 360 & - & 143 & 159 & - & 17 & 18 \\
                               &                       & 960 x 540 & - & 90 & 84 & - & 9 & 10 \\
                               & \multirow{3}{*}{FP32} & 480 x 270 & - & 130 & 125 & - & 15 & 21 \\
                               &                       & 640 x 360 & - & 95 & 95 & - & 8 & 14 \\
                               &                       & 960 x 540 & - & 40 & 50 & - & 3 & 7 \\ \hline
\end{tabular}
% }
\caption{Computational efficiency benchmark across various HW systems for different input sizes and data types.}
\label{tb:benchmark}
\end{table*}

\begin{figure}[h]
    \centering
    \includegraphics[width=1\columnwidth]{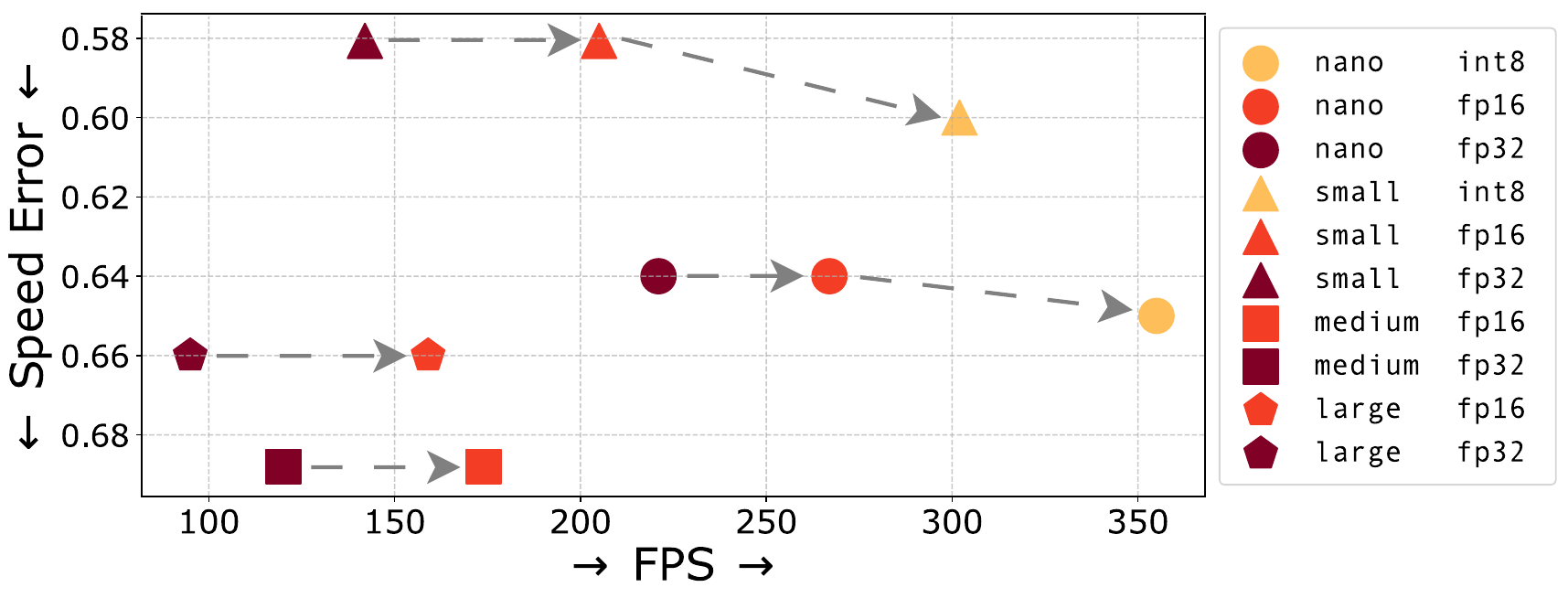}
    \caption{Quantization increases performance of all model configurations without a significant loss in median speed estimation accuracy, graph contains data for Titan V in $640 \times 360$ input resolution. }
    \label{img:quantization}
\end{figure}

% Table \ref{tb:benchmark} shows the FPS obtained on \TODO{asi tu treba napisat ten eval postup, teda ze ake data to boli} for different HW. The results show significant gains over previous SOTA method~\cite{kocur2020detection} especially for the smaller models. The quantized versions of the two best perorming models (Nano and Small with 640x360 input resolution) are capable of running real-time even on onlder edge device (Xavier).

When evaluating the computational efficiency of our speed estimation pipeline it is important to consider the traffic density. Increased traffic density translates into more detections per frame and thus increased computational load during the NMS stage of object detection. The three sessions from the test split (sessions 4-6) contain 19.28, 33.52 and 24.38 average cars per minute respectively. To benchmark the FPS across different hardware, we used 10 minutes of two videos (center and left view) from session 6 in the test set. With 24.38 average cars per minute, this session represents an average traffic density from the test split. For every video, the mean value of FPS was calculated. Finally, after processing both benchmark videos, the overall FPS was calculated as the mean value of the previously calculated mean for the two videos. The FPS estimates include the part of the computation performed on the CPU, which were carried it out in a multi-threaded manner to enable the optimal utilization of computational resources. Table \ref{tb:benchmark} shows the obtained FPS on for the tested HW. The results show significant gains over previous SOTA method~\cite{kocur2020detection} especially for the smaller models. The quantized versions of the two best performing models (Nano and Small with 640x360 input resolution) are capable of running real-time even on an older edge device (Xavier). We also show that on powerful hardware it is possible to process multiple videos simultaneously. 

\begin{figure}[h]
     \centering
     \includegraphics[width=1\columnwidth]{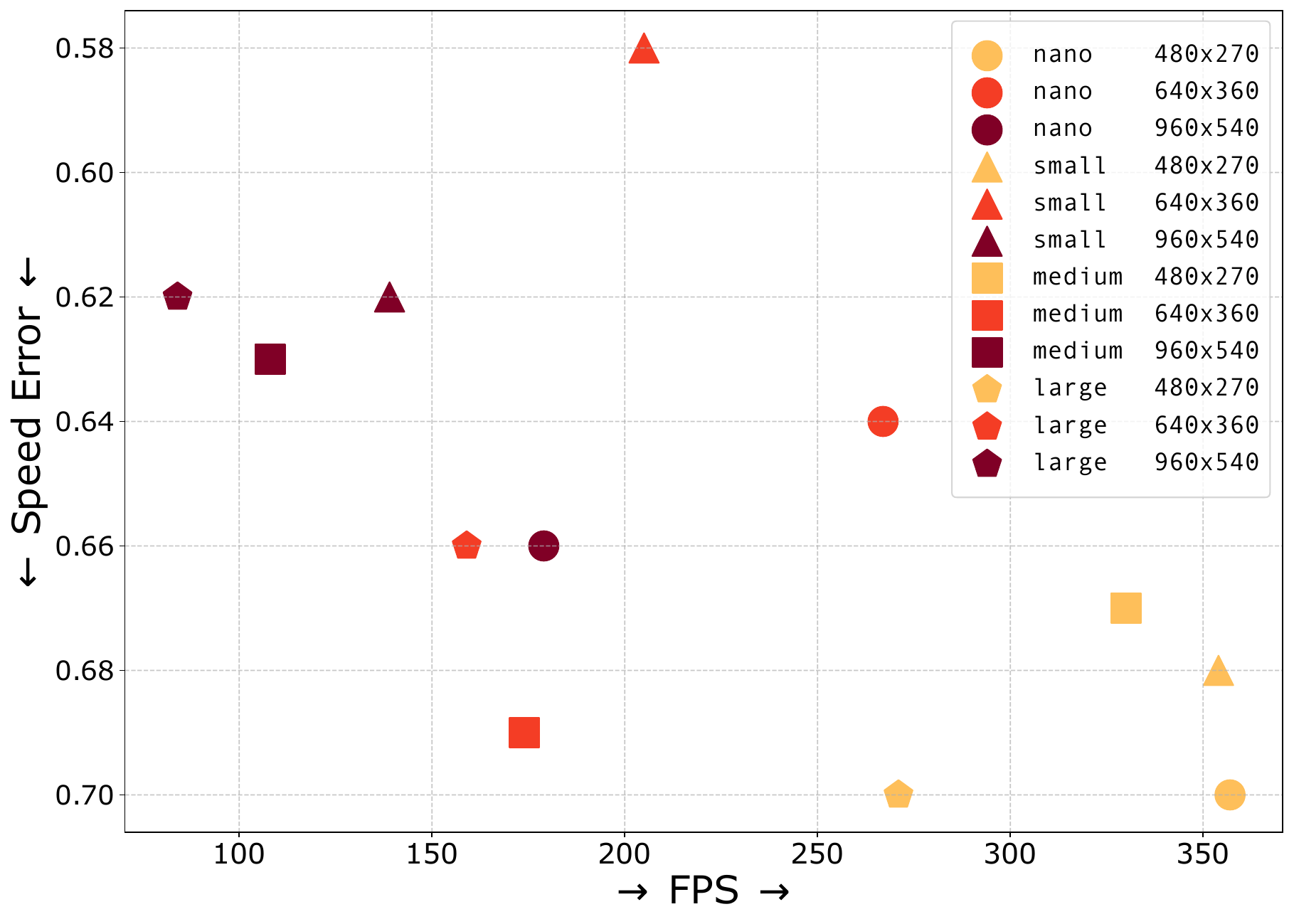}
     \caption{Comparison of median speed measurement error and performance for different model and input sizes on Titan V. Input size variations are represented by colors, clearly illustrating the trade-off between speed and error in this dimension. Difference in model size is denoted with shape and the trend is not so predictable, with Small model achieving the lowest error.}
     \label{img:comp}
\end{figure}

When considering the speed-vs-accuracy trade-off our evaluation shows the clear superiority of smaller models for the vehicle speed estimation task. Furthermore, quantization and computation in the decreased precision greatly improve computational speed with only a minor decrease in speed measurement accuracy (see Figure~\ref{img:quantization}).
Lowering the input size is also very predictable way of increasing the performance as visualized in Figure~\ref{img:comp}. 
%The middle input resolution of 640x360 in combination with Small model with smaller receptive field achieved the best generalization on the test set. 

The Nano model, particularly in its quantized form and on smaller inputs, achieves a very high performance of around 400FPS on desktop GPUs. Based on our experiments, it is not possible to push this performance much further even with stronger hardware, as it is already constrained by the speed of memory access. Therefore, we recommend the Nano model primarily on weaker hardware, such as Xavier. In all other situations, we advise the usage of the Small model, which has demonstrated the best overall performance. 

\begin{table}[]
\begin{center}
% \resizebox{0.8\linewidth}{!}{
\small
\begin{tabular}{cccc}
  Model &
  \begin{tabular}[c]{@{}c@{}}Input size \\ (px)\end{tabular} &
  \begin{tabular}[c]{@{}c@{}}Low Density \\ (FPS)\end{tabular} &
  %\begin{tabular}[c]{@{}c@{}}Medium Traffic Titan V\\ Session 6 (FPS)\end{tabular} &
  \begin{tabular}[c]{@{}c@{}}High Density\\ (FPS)\end{tabular} 
  \\ \hline
\multirow{3}{*}{Nano}          & 480 x 270 & 353 & 319 \\
                               & 640 x 360 & 297 & 258 \\
                               & 960 x 540 & 185 & 175 \\ \hline 
\multirow{3}{*}{Small}         & 480 x 270 & 351 & 313 \\
                               & 640 x 360 & 260 & 239 \\
                               & 960 x 540 & 156 & 145 \\ \hline 
\multirow{3}{*}{Medium}        & 480 x 270 & 288 & 262 \\
                               & 640 x 360 & 197 & 184 \\
                               & 960 x 540 & 114 & 108 \\ \hline 
\multirow{3}{*}{Large}         & 480 x 270 & 249 & 229 \\
                               & 640 x 360 & 168 & 159 \\
                               & 960 x 540 & 89 & 87 \\ \hline 
\end{tabular}
% }
\end{center}
\caption{Speed benchmark for different traffic situations. To evaluate computational demands we selected one minute segments from the test set during which 7 and 33 vehicles passed under the cameras representing low and high traffic density respectively. The results are shown for the Titan V system.}
\label{tb:situations}
\end{table}

We also investigate the effect of traffic density on computational demands. We benchmarked our models in two different situations. We picked a low-density situation by using a 1 minute long sequence from the test sessions with just 7 cars passing in the view of the camera. On the other hand, a different 1 minute long sequence with 33 cars was used to represent a situation with high traffic density. See Table~\ref{tb:situations} for comparison of FPS in these two situations. We provide data only for Titan V, but results for other systems are similar. In the dense traffic conditions the FPS rate drops noticeably, but not more than by 10\%.

\section{Discussion}

In this paper we have presented a method for vehicle speed estimation by improving the previous state-of-the-art method~\cite{kocur2020detection}. We have presented multiple variants of the proposed method with different model sizes and perform extensive evaluation of the real-world computational requirements using a range of hardware options from edge devices to desktop-grade hardware. Our expected use-case of the system is on locally deployed devices, with emphasis on edge devices. Server-grade accelerators were therefore not considered. Our evaluation shows that the presented improvements lead to significantly lower computational demands compared to the previous method while achieving similar or better vehicle speed measurement accuracy and better vehicle detection precision and recall.

Our evaluation also shows that minor improvements in bounding box localization accuracy do not necessarily translate to improved vehicle speed estimation accuracy. This may be explained by worse generalization capability of large models and by the fact that speed estimate is aggregated over multiple frames. Due to this, minor improvements in bounding box localization do not necessarily improve the speed estimate. We also show that computation using lower precision and post-training-quantization greatly benefit computational efficiency while reducing speed estimation accuracy only marginally, thus making them a clear choice for deploying vision-based vehicle speed estimation systems in practice.

\begin{acknowledgements}
Funded by the EU NextGenerationEU through the Recovery and Resilience Plan for Slovakia under the project No. ``InnovAIte Slovakia, Illuminating Pathways for AI-Driven Breakthroughs'' No.~09I02-03-V01-00029.
\end{acknowledgements}



\bibliographystyle{spmpsci}      % mathematics and physical sciences
\bibliography{references.bib}   % name your BibTeX data base

%\begin{thebibliography}{}
% etc
%\end{thebibliography}

% \input{parts/appendix.tex}

\end{document}

\begin{table*}[]
\centering
% \small
\resizebox{\textwidth}{!}{
\begin{tabular}{lcccccc}
Model &
  \begin{tabular}[c]{@{}c@{}}Input size \\ (px)\end{tabular} &
  \begin{tabular}[c]{@{}c@{}}Mean error \\ (km/h)\end{tabular} &
  \begin{tabular}[c]{@{}c@{}}Median error \\ (km/h)\end{tabular} &
  \begin{tabular}[c]{@{}c@{}}95-th percentile \\ (km/h)\end{tabular} &
  \begin{tabular}[c]{@{}c@{}}Mean precision \\ (\%)\end{tabular} &
  \begin{tabular}[c]{@{}c@{}}Mean recall \\ (\%)\end{tabular} \\ \hline
\multirow{3}{*}{Nano}          & 480 x 270 & 0.87          & 0.70          & 2.09          & 88.08          & 91.23           \\
                               & 640 x 360 & 0.80          & 0.64          & 2.04          & 89.69          & 92.25           \\
                               & 960 x 540 & 0.85          & 0.66          & 2.17          & 90.60          & \textbf{92.74}  \\ \hline
\multirow{3}{*}{Nano distill}  & 480 x 270 & 0.98          & 0.73          & 2.09          & 88.08          & 91.23           \\
                               & 640 x 360 & 0.90          & 0.73          & 2.24          & 86.00          & 89.30           \\
                               & 960 x 540 & 0.87          & 0.70          & 2.19          & 84.73          & 86.92           \\ \hline
\multirow{3}{*}{Small}         & 480 x 270 & 0.87          & 0.68          & 2.16          & 91.08          & 91.85           \\
                               & 640 x 360 & \textbf{0.76} & \textbf{0.58} & \textbf{2.01} & 91.02          & 92.16           \\
                               & 960 x 540 & 0.81          & 0.62          & 2.08          & \textbf{92.09} & 91.86           \\ \hline
\multirow{3}{*}{Small distill} & 480 x 270 & 0.91          & 0.73          & 2.24          & 85.45          & 91.00           \\
                               & 640 x 360 & 0.84          & 0.65          & 2.16          & 85.05          & 89.68           \\
                               & 960 x 540 & 0.83          & 0.70          & 2.16          & 84.73          & 90.07           \\ \hline
\multirow{3}{*}{Medium}        & 480 x 270 & 0.84          & 0.67          & 2.15          & 91.55          & 87.44           \\
                               & 640 x 360 & 0.86          & 0.69          & 2.21          & 91.32          & 90.68           \\
                               & 960 x 540 & 0.82          & 0.63          & 2.13          & 91.23          & 91.01           \\ \hline
\multirow{3}{*}{Large}         & 480 x 270 & 0.87          & 0.70          & 2.25          & 90.85          & 91.12           \\
                               & 640 x 360 & 0.84          & 0.66          & 2.16          & 90.34          & 90.67           \\
                               & 960 x 540 & 0.81          & 0.62          & 2.14          & 91.11          & 89.61           \\ \hline
\end{tabular}
}
\caption{Speed measurement evaluation on BrnoCompSpeed split test C, using FP16.}
\label{tb:fp16titan}
\end{table*}